\documentclass{article}
\usepackage{ijcai22}

\usepackage{times}

\usepackage{soul}
\usepackage{url}
\usepackage[hidelinks]{hyperref}
\usepackage[utf8]{inputenc}
\usepackage[small]{caption}
\usepackage{graphicx}
\usepackage{amsmath}
\usepackage{booktabs}

\usepackage{latexsym}
\usepackage{float} 
\usepackage{subfigure} 
\usepackage{algorithm}
\usepackage{algorithmic}

\usepackage[T1]{fontenc}
\usepackage{microtype}

\title{A Masked Image Reconstruction Network for Document-level Relation Extraction}

\author {
    Liang Zhang\textsuperscript{\rm 1} 
    Yidong Cheng \textsuperscript{\rm 1}\\
    \textsuperscript{\rm 1} Department of Artificial Intelligence, School of Informatics, Xiamen University\\
    \texttt{ \{lzhang,ydcheng\}@stu.xmu.edu.cn} \\
}

\begin{document}

\maketitle

\begin{abstract}
Document-level relation extraction aims to extract relations among entities within a document. 
Compared with its sentence-level counterpart, Document-level relation extraction requires inference over multiple sentences to extract complex relational triples. 
Previous research normally complete reasoning through information propagation on the mention-level or entity-level document-graphs, regardless of the correlations between the relationships. 
In this paper, we propose a novel \textbf{D}ocument-level \textbf{R}elation \textbf{E}xtraction model based on a \textbf{M}asked \textbf{I}mage \textbf{R}econstruction network (\textbf{DRE-MIR}), which models inference as a masked image reconstruction problem to capture the correlations between relationships. 
Specifically, we first leverage an encoder module to get the features of entities and construct the entity-pair matrix based on the features. 
After that, we look on the entity-pair matrix as an image and then randomly mask it and restore it through an inference module to capture the correlations between the relationships. 
We evaluate our model on three public document-level relation extraction datasets, i.e. DocRED, CDR, and GDA. 
Experimental results demonstrate that our model achieves state-of-the-art performance on these three datasets and has excellent robustness against the noises during the inference process.

\end{abstract}

\section{Introduction}
Relation extraction (RE) aims to identify the semantic relations between entities from raw texts, which is of great importance to many real-world applications \cite{c:1,c:2,c:3}.
Previous researches focused on sentence-level RE, which predicts the relationship between entities in a single sentence \cite{c:5,c:4,c:7}. 
However, large amounts of relationships are expressed by multiple sentences in real life \cite{c:8,c:9}.
Therefore, many recent works have made efforts to extend sentence-level RE to document-level RE \cite{c:8,c:10,c:11,c:12,c:13}.

\begin{figure}[t]
\centering
\includegraphics[width=0.95 \columnwidth]{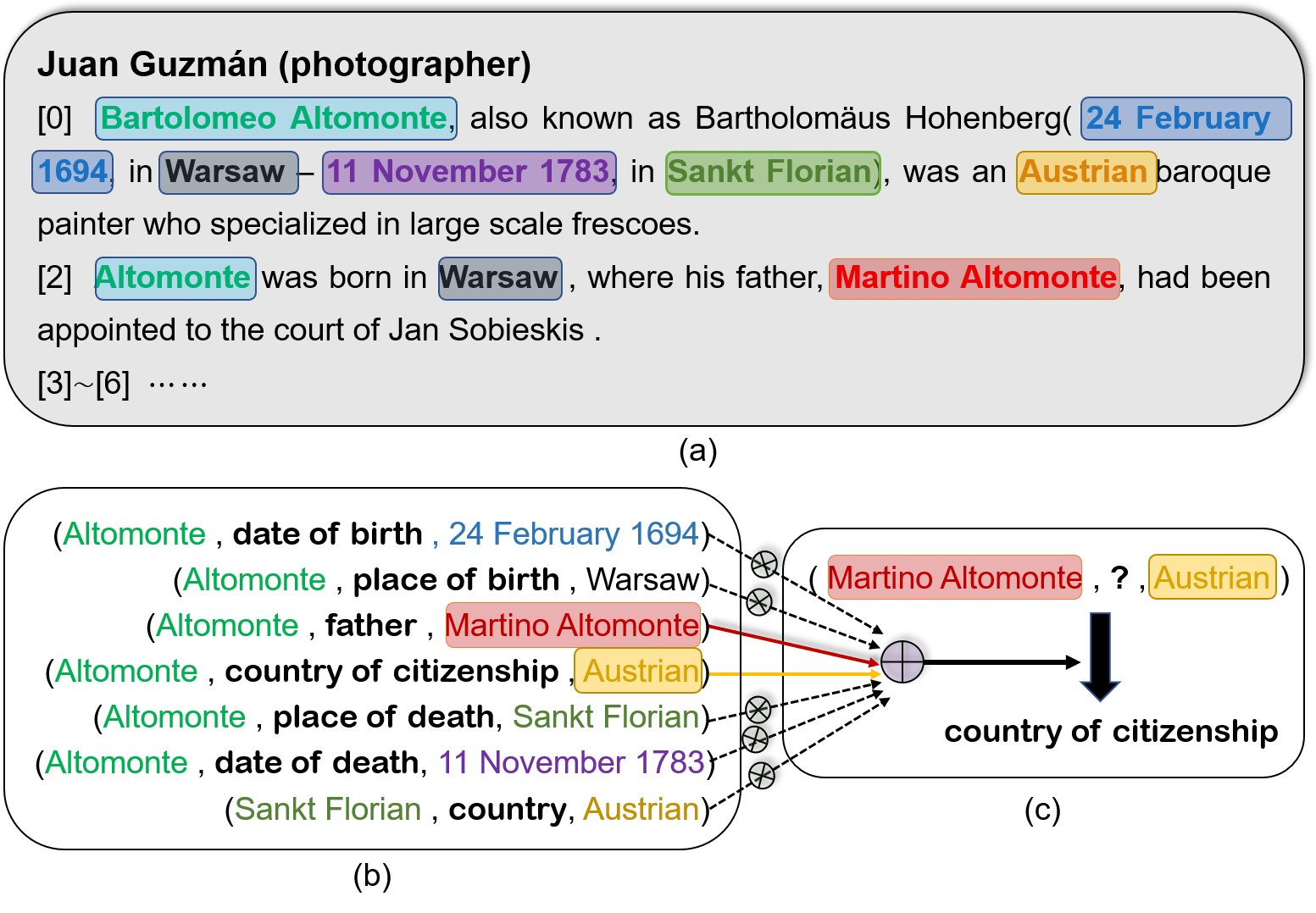} 
\caption{An example comes from the DocRED dataset, which shows that the use of correlation between relations (triple) to infer complex inter-sentence relations. (a) is a document, in which different colors represent different entities. (b) lists some intra-sentence relations, which can be easily identified. (c) shows an  inter-sentence relations which require reasoning techniques to be identified. The arrows between (b) and (c) indicate the correlation among relations.}
\label{fig1}
\end{figure}

\begin{figure*}[t]
\centering
\includegraphics[width=0.80 \textwidth]{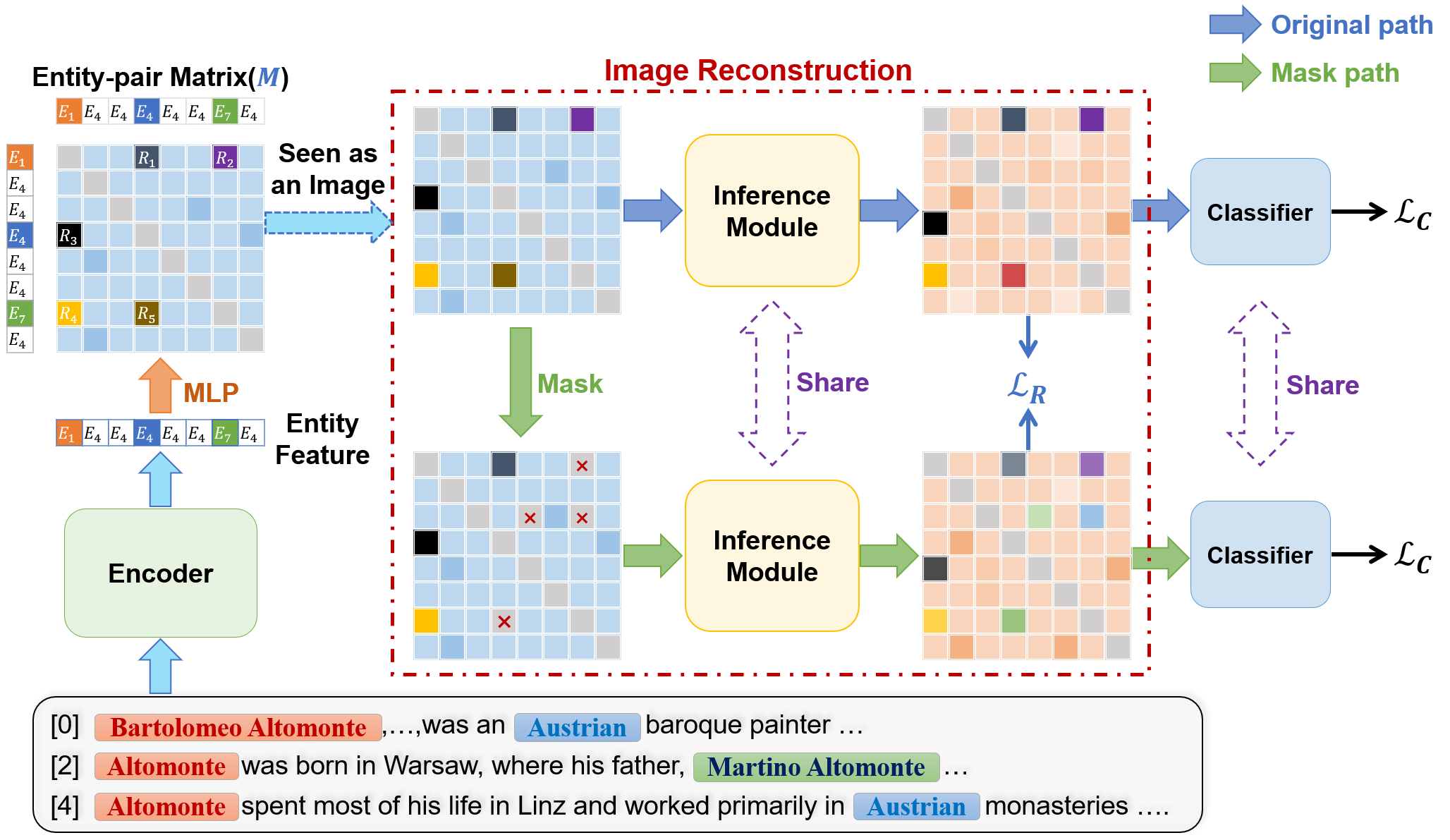} 
\caption{The overall architecture of our DRE-MIR model. 
Firstly, the Encoder encodes the input document to obtain the entities embedding ($E_s$,$E_o$), and then we obtain the Entity-pair Matrix $M$ through the linear layer.
Secondly, we treat the entity-pair matrix $M$ as an image, and then randomly mask it and
restore it through an inference module. 
Through the Masked Image Reconstruction (MIR) task, our inference module can learn how to use the correlation between relationships to infer the masked relationship.
Moreover, the MIR task contains two paths, i.e. the Original path and the Mask path.
Finally, we utilize a classifier to predict the relationship of each entity pair.
$L_R$ and $L_C$ represent reconstruction loss and classification loss, respectively.
}
\label{fig3}
\end{figure*}

Compared with sentence-level RE where a sentence contains only one entity pair to be classified, document-level RE requires the model to classify the relations of multiple entity pairs simultaneously and the entities involved in a relationship may appear in different sentences.
Besides, the document-level RE also poses a great challenge, i.e. relation inference. 
As shown in Figure~\ref{fig1}, it is easy to identify the intra-sentence relations shown in Figure 1b, such as (\textit{Altomonte, date of birth, 24 February 1694}), (\textit{Altomonte, father, Martino Altomonte}), and (\textit{Altomonte, country of citizenship, Austrian}), owing to two related entities appear in the same sentence.
However, it is more challenging for a model to predict the inter-sentential relations between \textit{Martino Altomonte} and \textit{Austrian} because the document does not explicitly express the relationship between them.
This type of inter-sentential relations can only be identified through reasoning techniques.
According to the statistics of the DocRED \cite{c:8} dataset which is a well-known document-level RE dataset, Most of the relation instances (61.1\%) require reasoning to be identified in document-level RE, which indicates that reasoning is essential for the document-level RE.

To extract such complex inter-sentence relations, most current approaches constructed a document-level graph based on heuristics, structured attention, or dependency structures \cite{c:10,c:14,c:15,c:16}, and then perform inference with graph convolutional network (GCN) \cite{c:17,c:18} on the document-level graph.
It should be noted that methods of this type complete reasoning through the information transfering between mentions or entities.
Meanwhile, considering the transformer architecture can implicitly model long-distance dependencies and can be regarded as a token-level fully connected graph, some studies \cite{c:19,c:11} implicitly infers through the pre-trained model rather than via the document-level graphs.

However, these methods ignore the correlation between relationships.
As shown in Figure~\ref{fig1}, we can easily infer the inter-sentence relation (\textit{Martino Altomonte, country of citizenship, Austrian}) through the correlation between the relationships.
Specifically, the model needs to firstly capture the correlation among (\textit{Altomonte, father, Martino Altomonte}), (\textit{Altomonte, country of citizenship, Austrian}), and (\textit{Martino Altomonte, country of citizenship, Austrian}), and then use reasoning techniques to identify this complex inter-sentential relation as shown in Figure 1c.

To capture the interdependencies among the multiple relationships, DocuNet \cite{c:13} formulates the document-level RE as a semantic segmentation problem and uses a U-shaped segmentation module over the image-style feature map to capture global interdependencies among triples.
The DocuNet model has achieved the latest state-of-the-art performance, which shows that the correlation between relationships is essential for the document-level RE.
However, capturing correlations between relations through convolutional neural networks is unintuitive and inefficient due to the intrinsic distinction between entity-pair matrices and image.

In this paper, we followed the DocuNet and model the document-level RE as a table filling problem.
We first construct an entity-pair matrix, where each point represents the relevant feature of an entity pair.
Then, the document-level RE model labels each point of the entity-pair matrix with the corresponding relationships class.
Meanwhile, we also treat the entity-pair matrix as an image.
To more effectively capture the interdependencies among the relations, we propose a novel \textbf{D}ocument-level \textbf{R}elation \textbf{E}xtraction model based on a \textbf{M}asked \textbf{I}mage \textbf{R}econstruction network (\textbf{DRE-MIR}), which formulates the inference problem in document-level RE as a masked image reconstruction problem.
As shown in Figure~\ref{fig3}, we first randomly mask the entity-pair matrix, and then reconstruct the entity pair matrix through the inference model.
Through this the Masked Image Reconstruction \textbf{(MIR)} task, our model can learn how to infer masked points with the help of correlations between relations.
Moreover, to more efficiently and intuitively reconstruct the masked points in the entity-pair matrix, we propose an Inference Multi-head Self-Attention (\textbf{I-MSA}) module which can greatly improve the inference ability of the model.
As shown in Figure~\ref{fig2}, the I-MSA contains four heads and each head corresponds to an inference mode including: \textbf{A$\rightarrow$${*}$ + ${*}$$\rightarrow$B$\implies$A$\rightarrow$B}, \textbf{A$\rightarrow$${*}$ + B$\rightarrow$${*}$$\implies$A$\rightarrow$B}, \textbf{${*}$$\rightarrow$A + B$\rightarrow$${*}$$\implies$A$\rightarrow$B}, and  \textbf{${*}$$\rightarrow$A + ${*}$$\rightarrow$B$\implies$A$\rightarrow$B}. 

Our contributions can be summarized as follows:
\begin{itemize}
\item To the best of our knowledge, our method is the first approach that treat the inference problem in document-level RE as an image reconstruction problem.
\item We introduce the I-MSA to improve the model's ability to reconstruct the masked entity-pair matrix.
\item Experimental results on three public document-level RE datasets shows that our Dense-CCNet model can achieve state-of-the-art performance.
\end{itemize}

\section{Method}
In this section, we introduce in detail our DRE-MIR model. 
As shown in Figure~\ref{fig3}, the DRE-MIR mainly consists of 3 modules, i.e. encoder
module, inference module, and classifier module.
We first describe the encoder module in Section~\ref{sec2.1}, then introduce the core module, i.e. inference module, in Section~\ref{sec2.2} , finally we describe our classifier module and loss function in Section~\ref{sec2.3}.

\subsection{Encoder Module}
\label{sec2.1}
\cite{c:20} and \cite{c:21} verified that marking entities in the input sentence by entity type can effectively improve the performance of sentence-level RE model.
However, in document-level RE, each entity has multiple mentions and it is important to gather all the mention information for each entity.
Therefore, we use the entity type and entity id to mark the mentions in the document, which not only can incorporate the entity type information earlier but also help to improve the aggregation of the mention information.
Specifically, given document $D=\{w_i\}_{j=1}^{l}$ containing $l$ words, we first mark the mention in the document by inserting special symbols $\left \langle e_{t} \right \rangle$ and $\left \langle e_{id} \right \rangle$ at the start and end position of the mentions, where $e_{t}$ and $e_{id}$ respectively represent the entity type and entity id of the mention.
Then we feed the adapted document to the pre-trained language model to obtain the context embedding of each word in the document:
\begin{equation}
H=[h_1,...,h_l] = Encoder([w_1,...,w_l]) .
\end{equation}
Finally, we utilize the average of the embeddings of $\left \langle e_{t} \right \rangle$ and $\left \langle e_{id} \right \rangle$ to represent  the mention.

For an entity $e_i$ with mentions $\{m_j^i\}_j^{N_{e_i}}$, we follow \cite{c:11} and \cite{c:13}, and leverage logsumexp pooling \cite{c:22}, a smooth version of max pooling, to obtain the embedding $h_{e_i}$ of entity $e_i$:
\begin{equation}
h_{e_i}=\log\sum_{j=1}^{N_{e_i}}exp(h_{m_j^i}) .
\end{equation}
In addition, we calculate an entity-pair-aware context representation $c_{s,o}$ for each entity pair $(e_s,e_o)$, which represents the contextual information in the document that the entity $e_s$ and the entity $e_o$ together pay attention to.
The $c_{s,o}$ is formulated as:
\begin{equation}
	\begin{split}
	    c_{s,o}&=Ha_{s,o}   ,   \\
        a_{s,o}&=softmax(A_s*A_o) ,
	\end{split}
\end{equation}
where $A_s$($A_o$) refers to the attention score that entity $e_s$($e_o$) pays attention to each word in the document, $H$ is the document embedding, and $*$ refers to element-wise multiplication.

Finally, we construct an entity-pair matrix $M{\in}R^{N_e\times N_e\times d}$ as follows:
\begin{equation}
	\begin{split}
	    M_{s,o}&=FFN([u_s,u_o]) ,   \\
	    u_s&=W_s[h_{e_s},h_{doc},c_{s,o}] , \\
	    u_o&=W_o[h_{e_o},h_{doc},c_{s,o}] ,
	\end{split}
\end{equation}
where $N_e$ represents the number of entities, $FFN()$ refers to a feed-forward neural network, $W_o$ and $W_s$ are the learnable weight matrix, $h_{doc}$ is [CLS] token embedding which is used to represent the information of the entire document.

\subsection{Inference Module}
\label{sec2.2}
After getting the entity-pair matrix, we treat it as an image.
We obtain a masked image by randomly masking the pixels of the original image and reconstruct the masked image through an inference module, as shown in Figure~\ref{fig3}.
Through this the Masked Image Reconstruction \textbf{(MIR)} task, our inference module can learn how to infer the masked pixels from the unmasked pixels by the correlation between the relationships.

\begin{figure}[t]
\centering
\includegraphics[width=0.95 \columnwidth]{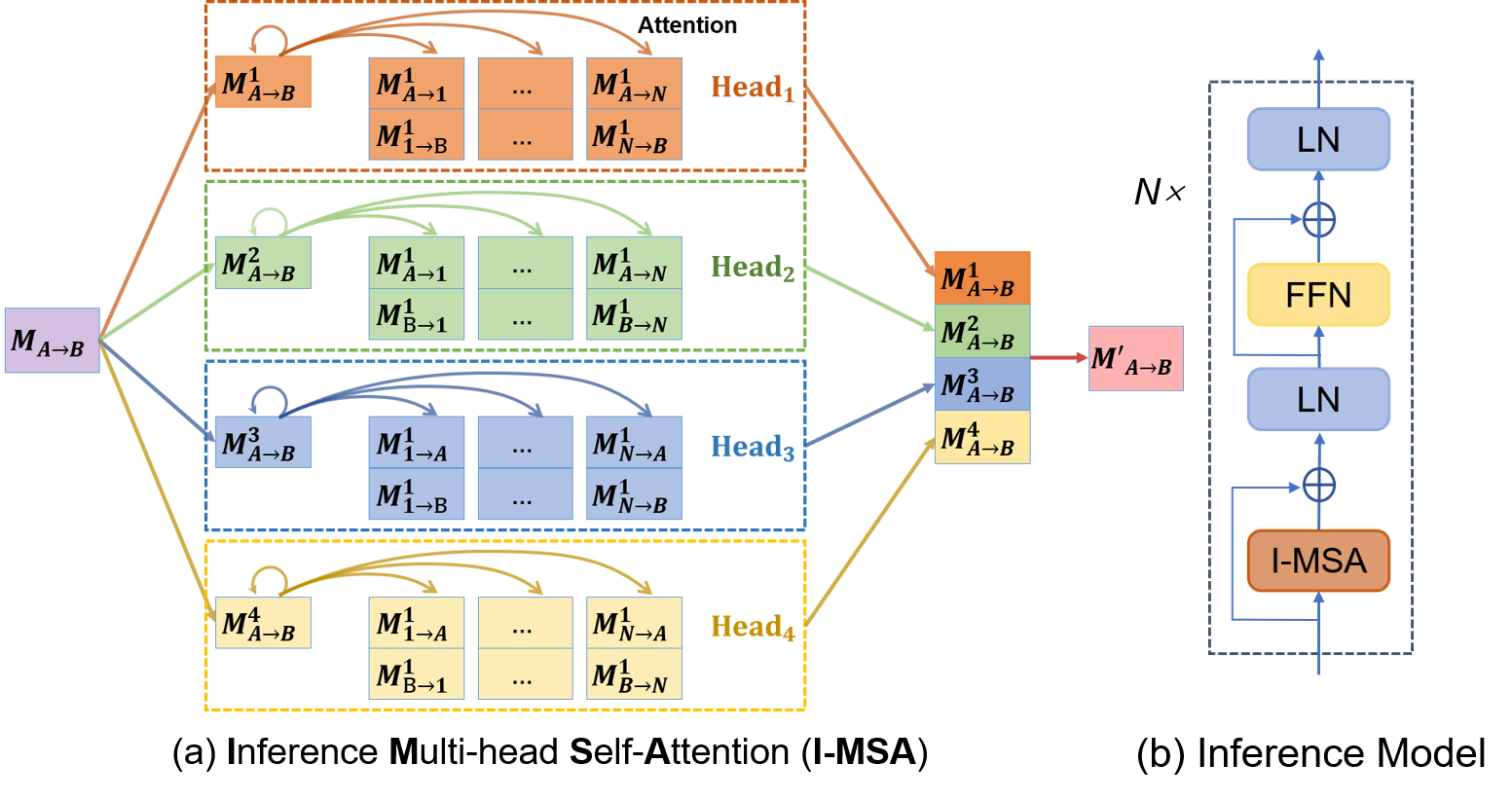} 
\caption{
(a) The architecture of the Inference Multi-head Self-Attention (I-MSA), which is a variant of multi-head self-attention (MSA). 
The I-MSA has four types of heads and each head corresponds to one inference mode.
(b) Inference module, which is a variant of Transformer's encoder by replacing MSA with I-MSA.
}
\label{fig2}
\end{figure}

Our inference module is a variant of Transformer's encoder, which replaces Multi-head Self-Attention (MSA) with Inference Multi-head Self-Attention (I-MSA), as shown in Figure~\ref{fig2}.
The I-MSA contains four heads and each head corresponds to an inference mode including: \textbf{A$\rightarrow$${*}$ + ${*}$$\rightarrow$B$\implies$A$\rightarrow$B}, \textbf{A$\rightarrow$${*}$ + B$\rightarrow$${*}$$\implies$A$\rightarrow$B}, \textbf{${*}$$\rightarrow$A + B$\rightarrow$${*}$$\implies$A$\rightarrow$B}, and  \textbf{${*}$$\rightarrow$A + ${*}$$\rightarrow$B$\implies$A$\rightarrow$B}. 
For example, for head 1 in Figure~\ref{fig2} which corresponds to \textbf{A$\rightarrow$${*}$ + ${*}$$\rightarrow$B$\implies$A$\rightarrow$B} inference mode, we first concatenate the corresponding pixels in the A-th row and B-th column of the image, $\{[M^1_{A\rightarrow1}, M^1_{1\rightarrow B}], \cdots, [M^1_{A\rightarrow N}, M^1_{N\rightarrow B}]\}$, and perform dimensionality reduction through a linear layer, $\{M^1_{A\rightarrow1\rightarrow B}, \cdots, M^1_{A\rightarrow N \rightarrow B}\}$.
Then, $M^1_{A,B}$ performs an attention operation on $\{M^1_{A\rightarrow1\rightarrow B}, \cdots, M^1_{A\rightarrow N \rightarrow B};M^1_{A,B}\}$.
The whole process can be formulated as follows:
\begin{equation*}
	\begin{split}
	    M^1_{A,B}=Attention(&M^1_{A,B}W^Q,M^1_{inf}W^K,M^1_{inf}W^V) ,  \\
	    M^1_{inf}=\{M^1_{A\rightarrow1\rightarrow B}&, \cdots, M^1_{A\rightarrow N \rightarrow B};M^1_{A,B}\} , \\
	    M^1_{A\rightarrow*\rightarrow B}=Liner(&[M^1_{A\rightarrow *}, M^1_{*\rightarrow B}]) ,
	\end{split}
\end{equation*}
where, $[~ ]$ represents the concatenation operation, $\{ \}$ refers to a set, $W^Q$, $W^K$, $W^V$ are the learnable weight matrix.

Inspired by \cite{c:23} and \cite{c:24}, we reconstruct the distribution of the pixels of the masked image on the label, $p(r|M_{A,B})$, instead of reconstructing the raw pixels, $M_{A,B}$.
The reason is that labels are more information-dense than pixels and $p(r|M_{A,B})$ is closer to our target task, relation classification.
In addition, we reconstruct each pixel on the masked image including the masked pixel and the unmasked pixel, which is similar to the \cite{c:25} method.
In this way, the convergence of the model can be accelerated and better performance can be obtained.
Specifically, the original image $M_{o}$ and the masked image $M_m$ are first sequentially input to the inference module and the classifier module to obtain the probability distributions $p(r|M_{o})$ and $p(r|M_{m})$. 
Then, we reconstruct the masked image by minimizing bidirectional KL-divergence between the two distributions of corresponding pixels in the original image and masked image.
Finally, our reconstruction loss function $L_R$ is formulated as follows:
\begin{equation}
	\begin{split}
	    L_R&=\frac{1}{2}\cdot ( D_{KL}(p(r|M_{o})\|p(r|M_{m}))\\
	    &+D_{KL}(p(r|M_{m})\|p(r|M_{o}))) .   \\
	\end{split}
\end{equation}

\subsection{Classifier Module}
\label{sec2.3}
Our classifier module is a single linear layer.
The original image $M_{o}$ and the masked image $M_{m}$ are respectively input to the inference module to obtain the corresponding corrected image, $M'_{o}$ and $M'_{m}$.
Then, the relation probability of each entity pair is obtained through a linear layer:
\begin{equation}
	\begin{split}
	    P(r|M'_{*})&=\sigma(W_r M'_{*}) ,\\
	    M'_{*}&=Inference(M_{*}) ,
	\end{split}
\end{equation}
where $*\in \{o,m\}$, and $W_r$ is model parameters.

To alleviate the problem of unbalanced relationship distribution, we use adaptive-thresholding loss \cite{c:11} as our classification loss function $L_C$, which learns an adaptive threshold for each sample.
Specifically, a $TH$ class is introduced to separate positive classes and negative classes: positive classes would have higher probabilities than $TH$, and negative classes would have lower probabilities than $TH$.
The adaptive-thresholding loss is formulated as follows:
\begin{equation}
	\begin{split}
	    L_C&=L_1+L_2 ,\\
	    L_1&={-}\sum_{r{\in}P_D} \log \left(\frac{\exp(logit_r)}{\sum_{r'{\in}\{P_D,TH\}} \exp(logit_{r'})}\right) ,\\
	    L_2&={-}\log \left(\frac{\exp(logit_{TH})}{\sum_{r'{\in}\{N_D,TH\}} \exp(logit_{r'})}\right) ,
	\end{split}
\end{equation}
where $P_D$ and $N_D$ are the positive classes set and negative classes set respectively.

The training objective is to minimize the loss function $L$, which is defined as follows:
$$L = \alpha L_R + \beta L_C$$
$ \alpha $ and $ \beta $ are hyperparameters and we simply set them to 1.

\section{Experiments}
\subsection{Datasets}
We conduct experiments on three document-level RE datasets to evaluate our DRE-MIR model. 
The statistics of the datasets could be found in Appendix~\ref{appendix-a}.
\begin{itemize}
\item \textbf{DocRED} \cite{c:8}:
DocRED is a large-scale human-annotated dataset for document-level RE, which constructed from Wikipedia and Wikidata. 
DocRED contains 96 types of relations, 132,275 entities, and 56,354 relationship triples in total. 
In DocRED, more than 40.7\% of relational facts can only be extracted from multiple sentences, and 61.1\% of relational triples require various reasoning skills. 
We follow the standard split of the dataset, 3,053 documents for training, 1,000 for development and, 1,000 for the test.
\item \textbf{CDR} \cite{c:26}:
The Chemical-Disease Reactions dataset (CDR) consists of 1,500 PubMed abstracts, which is equally divided into three sets for training, development, and testing. CDR is aimed to predict the binary interactions between Chemical and Disease concepts. 
\item \textbf{GDA} \cite{c:27}:
The Gene-Disease Associations dataset (GDA) is a large-scale biomedical dataset, which is constructed from MEDLINE abstracts by method of distant supervision.
GDA contains 29,192 documents as the training set and 1,000 as the test set. 
GDA is also a binary relation classification task that identifies Gene and Disease concepts interactions. We follow \cite{c:15} to divide the training set into two parts, 23,353 documents for training and 5,839 for validation.
\end{itemize}

\begin{table*}[th]
\centering
\begin{tabular}{lcccccc}
\toprule
Model                & \multicolumn{4}{c}{Dev}                  & \multicolumn{2}{c}{Test} \\
                     & Ign$F_1$         & $F_1$     & Intra-$F1$    & Inter-$F1$   & Ign$F_1$     & $F_1$  \\ [2pt]\toprule
GEDA-$\rm BERT_{base}$\cite{c:36}       & 54.52     & 56.16     & -             & -        & 53.71      & 55.74       \\
LSR-$\rm BERT_{base}$\cite{c:14}       & 52.43     & 59.00     &65.26          &52.05     & 56.97      & 59.05       \\
GLRE-$\rm BERT_{base}$\cite{c:16}       & -         & -         & -             & -        & 55.40       & 57.40        \\
HeterGSAN-$\rm BERT_{base}$\cite{c:10}  & 58.13     & 60.18     & -             & -        & 57.12      & 59.45       \\ 
GAIN-$\rm BERT_{base}$\cite{c:37}       & 59.14     & 61.22     &67.10          &53.90     & 59.00         & 61.24       \\
[2pt] \toprule
$\rm BERT_{base}$\cite{c:38}            & -         & 54.16     &61.61          &47.15      & -          & 53.20        \\
BERT-$\rm TS_{base}$\cite{c:38}         & -         & 54.42     &61.80          &47.28      & -          & 53.92       \\
HIN-$\rm BERT_{base}$\cite{c:19}        & 54.29     & 56.31     & -             & -         & 53.7       & 55.60        \\
Coref$\rm BERT_{base}$\cite{c:39}       & 55.32     & 57.51     & -             & -         & 54.54      & 56.96       \\
ATLOP-$\rm BERT_{base}$\cite{c:11}      & 59.22     & 61.09     & -             & -         & 59.31      & 61.30        \\
SIRE-BERT\cite{c:42}                    & 59.82     & 61.60     & 68.07         & 54.01     & 60.18     & 62.05       \\
DocuNet-$\rm BERT_{base}$\cite{c:13}    & 59.86     & 61.83     & -             & -         & 59.93      & 61.86       \\[2pt] \toprule
DRE-MIR-$\rm BERT_{base}$ & \textbf{60.85$\pm$0.10} & \textbf{62.81$\pm$0.13} & \textbf{68.67$\pm$0.11} & \textbf{56.09$\pm$0.21}   & \textbf{60.91} & \textbf{62.85} \\ \bottomrule

\end{tabular}
\caption{
    Results (\%) on the development and test set of the DocRED. The scores of all the baseline models come from ATLOP \protect\cite{c:11} and SIRE \protect\cite{c:42}. 
    And, the results on the test set are obtained by submitting to the official Codalab.
}
\label{tab1} 
\end{table*}

\begin{table}[]
\centering
\begin{tabular}{p{5.7cm}cc} 
\toprule
Model               & CDR          & GDA         \\ [2pt] \toprule
BRAN\cite{c:9}                & 62.1         & -           \\
EoG\cite{c:15}                & 63.6         & 81.5        \\
LSR\cite{c:14}                & 64.8         & 82.2        \\
DHG\cite{c:40}                & 65.9         & 83.1        \\
GLRE\cite{c:16}                & 68.5         & -           \\
SciBERT\cite{c:31}        & 65.1         & 82.5        \\
ATLOP-SciBERT\cite{c:11}   & 69.4         & 83.9        \\  
DocuNet-SciBERT\cite{c:13} & 76.3         & 85.3     \\[2pt] \toprule
DRE-MIR-SciBERT    & \textbf{76.8}     & \textbf{86.4}    \\
\bottomrule
\end{tabular}
\caption{\label{tab2}Results (\%) on the biomedical datasets CDR and GDA.
}
\end{table}

\subsection{Experimental Settings}
Our model was implemented based on PyTorch and Huggingface’s Trans-formers \cite{c:28}. 
We used cased BERT-base \cite{c:29} as the encoder on DocRED and SciBERT-base \cite{c:31} on CDR and GDA.
Our model is optimized with AdamW \cite{c:32} with a linear warmup \cite{c:33} for the first 6\% steps followed by a linear decay to 0.
By default, we randomly mask 20\% of points in the entity-pair matrix and set the number of layers in the inference module to 3.
All hyper-parameters are tuned on the development set, some of which are listed in Appendix~\ref{appendix-b}.

\subsection{Results on the DocRED Dataset}
On the DocRED Dataset, we choose the following two types of models as the baseline:
\begin{itemize}
    \item \textbf{Graph-based Models}: This type of method uses graph convolutional networks (GCN) \cite{c:18,c:34,c:35}  to complete inference on document-level graphs, including GEDA \cite{c:36}, LSR \cite{c:14}, GLRE \cite{c:16}, GAIN \cite{c:10}, and HeterGSAN \cite{c:37}.
    
    \item \textbf{Transformer-based Models}: These models directly use pre-trained language models for document-level RE without using graph structures, including $\rm BERT$ \cite{c:37}, BERT-Two-Step \cite{c:38}, HIN-BERT \cite{c:19}, $\rm CorefBERT$ \cite{c:39}, and ATLOP-BERT \cite{c:11}.
\end{itemize}
In addition, we also consider the DocuNet \cite{c:13} and SIRE \cite{c:42} in the comparison. 
The DocuNet formulates document-level RE as a semantic segmentation problem and get the latest SOTA results. 
While, the SIRE represents intra- and inter-sentential relations in different ways, and design a new form of logical reasoning.

We follow \cite{c:8} and use $F_1$ and Ign$F_1$ as evaluation metrics to evaluate the performance of a model, where Ign$F_1$ denotes the $F_1$ score excluding the relational facts that are shared by the training and dev/test sets. 
Comparing all baseline model, our DRE-MIR model outperforms the latest state-of-the-art models by \textbf{1.14/1.11} $F_1$/Ign$F_1$ on the dev set and \textbf{1.29/1.1} $F_1$/Ign$F1$ on the test set. This demonstrates that our model has an excellent overall performance. 
Besides, comparing the graph-based state-of-the-art model, the  DRE-MIR model outperforms the GAIN model by \textbf{1.74/1.83} $F_1$/Ign$F_1$ on the dev set and \textbf{2.11/2.03} $F_1$/Ign$F_1$ on the test set. 
This shows that our model has better reasoning ability than the previous graph-based models.

The same as \cite{c:14,c:37} , we report Intra-$F_1$ / Inter-$F_1$ scores in Table~\ref{tab1}, which only consider either intra- or inter-sentence relations respectively. 
Compared with Intra-$F_1$, Inter-$F_1$ can better reflect the reasoning ability of the model.
we can observe that our DRE-MIR model improved the Inter-$F_1$ score by \textbf{3.28} compared with the SIRE model.
The improvement on Inter-$F_1$ demonstrates that our MIR task and Inference module can greatly improve the inference ability of the model.
Moreover, the improvement on Inter-$F_1$ is greater than on intra-$F_1$, which shows that the performance improvement of DRE-MIR is mainly contributed by the improvement of inter-sentence relations.

\begin{table}[]
\centering
\setlength{\tabcolsep}{2.8mm}{
\begin{tabular}{lcc}
\toprule
Model                               & Ign$F_1$      & $F_1$     \\ [2pt] \toprule
DRE-MIR                       & \textbf{60.97} & \textbf{62.96} \\ 
Only Mask path                       &60.13 	       &  62.20       \\
Only reconsitution masked point   & 59.50 	&   61.53     \\
w/o MIR                       &59.31  	       & 61.35       \\
w/o Inference Model                       &58.56 	&  60.46      \\
w/o I-MSA                       &54.67 	&  56.22      \\ \bottomrule
\end{tabular}}
\caption{
Ablation study of the DRE-MIR model on the development set of the DocRED. 
\textbf{w/o Inference Model} and \textbf{w/o MIR} removes the Inference Model and the MIR task from our mode, respectively;
\textbf{w/o I-MSA} replaces the Inference Multi-head Self-Attention (I-MSA) with the multi-head self-attention (MSA);
\textbf{w/o Inference Model} is our base model.
}
\label{tab3} 
\end{table}

\subsection{Results on the Biomedical Datasets}
On the two biomedical datasets, CDR and GDA, we compared our model with a large number of baseline models and recent state-of-the-art models including BRAN \cite{c:9}, EoG \cite{c:15}, LSR \cite{c:14}, DHG \cite{c:40}, GLRE \cite{c:16}, SciBERT \cite{c:31}, ATLOP \cite{c:11}, and DocuNet \cite{c:13}.

Experiment results on two biomedical datasets are shown in Table~\ref{tab2}.
Our DRE-MIR model achieves \textbf{76.9} $F1$ on the CDR dataset, which slightly out performs the DocuNe model by \textbf{0.3} $F_1$.
There are three possible reasons: (1) The CDR contains only two types of relations, which indicates that the correlation between relations is weak.
(2) The CDR dataset contains very few annotated samples, making it difficult for our model to learn the underlying correlations. 
(3) The samples in the CDR dataset contain few entities, which leads to a small entity pair matrix and weakens the effectiveness of our MIR task.
Although the GDA dataset also has problems (1) and (3), it is a large-scale corpus and contains a large number of samples. Therefore our model achieves \textbf{86.4} $F1$ score on the GDA dataset, which improves
\textbf{1.1} $F1$ compared with the DocuNe model.
Since the MIR task is a pre-training task in the field of machine vision, more data is required to obtain better performance.


\begin{table}[]
\centering
\setlength{\tabcolsep}{6.5mm}{
\begin{tabular}{lcc}
\toprule
Layer-number                            & Ign$F_1$      & $F_1$     \\ [2pt] \toprule
1-Layer                           & 60.38        & 62.31       \\
2-Layer                          & \textbf{60.85}       & \textbf{62.81}       \\\bottomrule

\end{tabular}}
\caption{Performance of DRE-MIR with different numbers of layers in the inference module on the development set of DocRED.}
\label{tab4}
\end{table}

\subsection{Ablation Study}
We conducted an ablation experiment to validate the effectiveness of different components of our DRE-MIR model on the development set of the DocRED dataset.
The results are listed in Table~\ref{tab3}.

The \textbf{w/o MIR} removes the MIR task from our model and only contains the original path in the DRE-MIR model.
The \textbf{w/o MIR} achieves an F1 score of 61.35, which outperforms the \textbf{w/o Inference Model}, our base model, by \textbf{0.99} $F1$.
This shows that our inference module has a certain inference ability even without the MIR task.
However, the DRE-MIR model without the MIR task (\textbf{w/o MIR}) has a performance drop of \textbf{1.61} $F1$, which proves that the MIR task can well improve the inference ability of our inference module.

The \textbf{Only Mask path} removes the original path from our model and only contains the Mask path.
The \textbf{Only Mask path} is a variant of the DRE-MIR model, and its image reconstruction method is similar to \cite{c:23}.
The \textbf{Only Mask path} leads to a drop of \textbf{0.76} $F1$ in performance, which proves that the original image played a guiding role in the reconstruction process of the masked image to further improve the performance of the model.

As can be seen from \textbf{w/o I-MSA}, replacing the I-MSA with the MSA resulted in a huge performance drop of \textbf{6.74} $F1$.
This shows that our I-MSA can greatly improve the inference ability of the Transformer.
We also introduce an experiment where only the masked pixels are reconstructed, i.e. \textbf{Only reconsitution masked point}, and observe a performance drop of \textbf{1.43} $F1$.
The possible reason is that the masked pixels may affect the unmasked pixels through our inference module, but reconstructing all the pixels can effectively alleviate this negative impact.

Overall, our model improves our base model by \textbf{2.5} $F1$, which fully demonstrates that our inference module and MIR task can effectively improve the inference ability of the model.

\begin{figure}[t]
\centering
\includegraphics[width=0.90 \columnwidth]{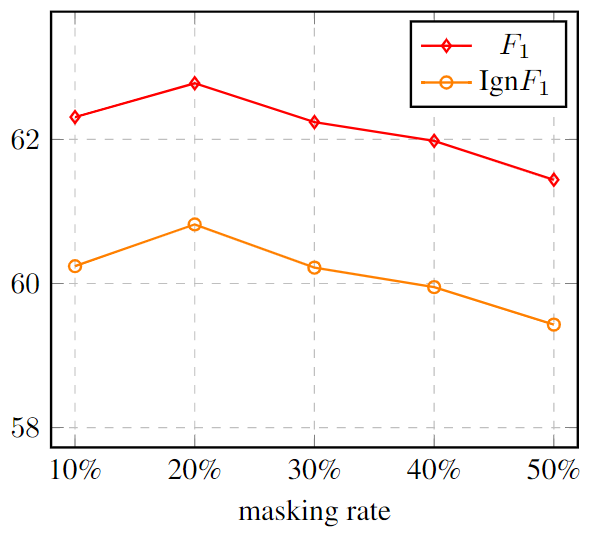}
\caption{Results of different masking rates used in the training process on the development set of DocRED. }
\label{fig4}
\end{figure}

\subsection{Analysis \& Discussion}
In this section, we will further discuss and analyze our DRE-MIR model from four aspects: (1) the number of layers in the inference module, (2) the masking rate during training, (3) the inference performance, and (4) the performance to restore the masked entity-pair matrix.

Table~\ref{tab4} shows the performance of the DRE-MIR model with different number of layers of inference modules.
We observe that increasing the number of layers from 1 to 2 improves the model performance by \textbf{1.07} $F_1$ score.
The possible reason is that increasing the number of layers can improve the multi-hop reasoning ability of the model.
However, the performance of the model is slightly improved by \textbf{0.21} $F_1$ when the number of layers is increased from 2 to 3.
Therefore, a two-layer inference module is sufficient for general cases.

Figure~\ref{fig4} shows that our model obtains the best performance when trained with a masking rate of 20\%.
However, our model still achieves a decent performance of \textbf{61.44/59.43} $F_1$/Ign$F_1$ when setting the masking rate to 50\%, which shows that our inference module has strong inference ability to restore the masked entity-pair matrix.
This also implies that using a larger masking rate to increase the training difficulty should achieve better performance under large-scale corpora, which is similar to the conclusions drawn from pre-training tasks in machine vision, such as MAE \cite{c:25}.

To evaluate the inference ability of the models, we follow \cite{c:37,c:42} and report Infer-$F_1$ scores in table~\ref{tab5}, which only considers relations that engaged in the relational reasoning process. 
We observe that our DRE-MIR model improves \textbf{2.71} Infer-$F_1$ compared with the GAIN model.
Removing the inference module from our model results in a performance drop of \textbf{4.10} Infer-$F_1$, which demonstrates that our inference module and the MIR task can improve the inference ability of the model.

\begin{figure}[t]
\centering
\includegraphics[width=0.95 \columnwidth]{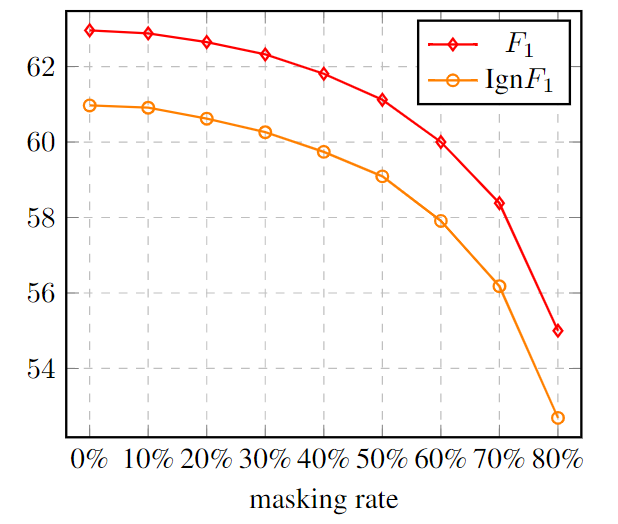} 
\caption{Results of different masking rates used the validating process on the development set of DocRED.}
\label{fig5}
\end{figure}

To evaluate the model's ability of restoring the masked entity-pair matrix, we also randomly mask the entity-pair matrix during validating.
We show the experimental results in Figure~\ref{fig5}.
Since we train our model with a masking rate of 20\%, the performance drop of the model is very slight when the masking rate is less than 20\%.
Our model has only a slight performance drop of \textbf{1.84/1.88} $F_1$/Ign$F_1$ with 50\% masking rate, which shows that our model has excellent robustness.
Even if the masking rate is increased to 80\%, our model still achieves a score of \textbf{55.00/52.69} $F_1$/Ign$F_1$, which is better than the BERT-$\rm TS_{base}$\cite{c:38} model.
This shows that our model has strong restoring ability.

\begin{table}[]
\centering
\begin{tabular}{p{3.5cm}ccc} 
\toprule
Model                       & Infer-$F_1$      & P         &R         \\ [2pt] \toprule
GAIN-GloVe             &40.82          & 32.76     &54.14           \\
SIRE-GloVe             & 42.72         & 34.83     &55.22        \\ \toprule
BERT-RE$_{base}$             & 39.62         & 34.12     &47.23        \\
RoBERTa-RE$_{base}$              & 41.78         & 37.97     &46.45        \\
GAIN-$\rm BERT_{base}$            & 46.89         & 38.71     &59.45           \\[2pt] \toprule
DRE-MIR-$\rm BERT_{base}$        &  \textbf{48.83}           &  \textbf{42.93}     & \textbf{56.61}    \\
w/o Inference Model          &46.70     &38.63    &59.06   \\
\bottomrule
\end{tabular}
\caption{Infer-$F_1$ results of the DRE-MIR model on the development set of DocRED. P: Precision, R: Recall.}
\label{tab5}
\end{table}

\section{Related Work}


Since many relational facts in real applications can only be recognized across sentences, a lot of recent work gradually shift their attention to document-level RE.
Due to graph neural network(GNN) can effectively model long-distance dependence and complete logical reasoning, Many methods based on document-graphs are widely used for document-level RE.
Specifically, they first constructed a graph structure from the document, and then applied the GCN \cite{c:18,c:51} to the graph to complete logical reasoning.
The graph-based method was first introduced by \cite{c:52} and has recently been extended by many works \cite{c:15,c:36,c:40,c:53,c:16,c:14,c:10,c:27}.
\cite{c:36} proposed the Graph Enhanced Dual Attention network (GEDA) model and used it to characterize the complex interaction between sentences and potential relation instances.
\cite{c:10} propose Graph Aggregation-and-Inference Network (GAIN) model.
GAIN first constructs a heterogeneous mention-level graph (hMG) to model complex interaction among different mentions across the document and then constructs an entity-level graph (EG), finally uses the path reasoning mechanism to infer relations between entities on EG.
\cite{c:14} proposed a novel LSR model, which constructs a latent document-level graph and completes logical reasoning on the graph.

In addition, due to the pre-trained language model based on the transformer architecture can implicitly model long-distance dependence and complete logical reasoning, some studies \cite{c:19,c:11,c:38} directly apply pre-trained model without introducing document graphs.
\cite{c:11} proposed an ATLOP model that consists of two parts: adaptive thresholding and localized context pooling, to solve the multi-label and multi-entity problems.
The SIRE \cite{c:42} represents intra- and inter-sentential relations in different ways, and design a new and straightforward form of logical reasoning.
Recently, the state-of-the-ar model, DocuNet \cite{c:13},  formulates document-level RE as semantic segmentation task and capture  global information among relational triples through the U-shaped segmentation module \cite{c:41}.

Furthermore, our work is inspired by recent pre-training research in the field of machine vision, such as BIET \cite{c:23}, IBOT \cite{c:24}, and MAE \cite{c:25}.
BEIT followed BERT \cite{c:29} and proposed a masked image modeling (MIM) task and a tokenizer to pre-train vision Transformers.
The tokenizer “tokenize” the image to discrete visual tokens, which is obtained by the latent codes of discrete VAE \cite{c:55}.
The IBOT can perform the MIM task with an online tokenizer and formulates the MIM task as knowledge distillation (KD) distillation problem.
The MAE develops an asymmetric encoder-decoder architecture, with an encoder that operates only on the visible subset of patches (without mask tokens), along with a lightweight decoder that reconstructs the original image from the latent representation and mask tokens.

\section{Conclusion and Future Work}
In this work, We first formulate the inference problem in document-level RE as a Masked Image Reconstruction (MIR) problem. 
Then, we propose an Inference Multi-head Self-Attention (I-MSA) module to restore masked images more efficiently.
The MIR task and the I-MSA module greatly improve the inference ability of our model.
Experiments on three public document-level RE datasets demonstrate that our DRE-MIR model achieved better results than the existing state-of-the-art model.
In the future, we will try to use our model for other inter-sentence or document-level tasks, such as cross-sentence collective event detection.

\bibliography{ijcai22}
\bibliographystyle{ijcai22}

\clearpage
\appendix
\begin{table}[htb]
\centering
\setlength{\tabcolsep}{2.5mm}{
\begin{tabular}{lccc}
\toprule
Dataset                     & DocRED     &CDR       &GDA  \\  [2pt] \toprule
Train                     &3053	    &500	    &23353         \\ 
Dev                       &1000	    &500	    &5839       \\
Test                      &1000	    &500	    &1000       \\
Relations                 &97	    &2	    &2        \\
Entities per Doc        &19.5	    &7.6	    &5.4        \\
Mentions per Doc        &26.2	    &19.2	    &18.5        \\ 
Entities per Sent        &3.58	    &2.48	    &2.28        \\     \bottomrule

\end{tabular}}
\caption{ Summary of DocRED, CDR and GDA datasets.
}
\label{tab6}
\end{table}

\begin{table}[htb]
\centering
\setlength{\tabcolsep}{2.5mm}{
\begin{tabular}{lccc}
\toprule
Hyperparam                     & DocRED     &CDR       &GDA  \\ 
                            & BERT     &SciBERT       &SciBERT  \\  [2pt] \toprule
Batch size                    &8	    &16	    &16         \\ 
Epoch                       &100	    &20	    &5 \\
lr for encoder                 &2e-5	    &2e-5	    &2e-5        \\
lr for other parts       &1e-4	    &5e-5	    &5e-5         \\
\bottomrule
\end{tabular}}
\caption{\label{tab7} Hyper-parameters Setting.}
\end{table}

\section{Datasets}
\label{appendix-a}

Table~\ref{tab5} details the statistics of the three document-level relational extraction datasets, DocRED, CDR, and GDA. These statis-tics further demonstrate the complexity of entity structure in document-level relation extraction tasks.

\section{Hyper-parameters Setting}
\label{appendix-b}
Table~\ref{tab6} details our hyper-parameters setting. All of our hyperparameters were tuned on the development set.

\end{document}